\documentclass[10pt,twocolumn,letterpaper]{article}

\usepackage{iccv}
\usepackage{times}
\usepackage{epsfig}
\usepackage{graphicx}
\usepackage{amsmath}
\usepackage{amssymb}

\usepackage{calc}

\usepackage{times}
\usepackage{amsfonts}

\usepackage{pifont}
\newcommand{\cmark}{\ding{51}}%
\newcommand{\xmark}{\ding{55}}%

\usepackage{epsfig}
\usepackage{graphicx}
\usepackage[dvipsnames,table]{xcolor}        

\usepackage[normalem]{ulem}
\usepackage{booktabs}
\usepackage{siunitx}

\usepackage{etoolbox}
\robustify{\bfseries}
\robustify{\itshape}

\usepackage{amsmath}
\usepackage{amssymb}
\usepackage{mathtools}

\usepackage{balance,flushend}

\usepackage[sort&compress,numbers]{natbib}

\makeatletter
\patchcmd{\NAT@test}{\else \NAT@nm}{\else \NAT@nmfmt{\NAT@nm}}{}{}

\DeclareRobustCommand\citepos
{\begingroup
  \let\NAT@nmfmt\NAT@posfmt
  \NAT@swafalse\let\NAT@ctype\z@\NAT@partrue
  \@ifstar{\NAT@fulltrue\NAT@citetp}{\NAT@fullfalse\NAT@citetp}}

\let\NAT@orig@nmfmt\NAT@nmfmt
\def\NAT@posfmt#1{\NAT@orig@nmfmt{#1's}}
\makeatother

\makeatletter
\def\NAT@spacechar{~}
\makeatother

\usepackage{siunitx}
\robustify\textbf
\robustify\bfseries
\usepackage{tikz}
\usetikzlibrary{
  calc,
  backgrounds,
  math,
  matrix,
  positioning,
  shapes.misc,
}

\usepackage{pgfplots}
\usepgfplotslibrary{
  colorbrewer,
}

\usepackage[pagebackref=true,breaklinks=true,letterpaper=true,colorlinks,bookmarks=false]{hyperref}
\hypersetup{
  colorlinks,
  linkcolor = BrickRed,
  citecolor = RoyalBlue,
  urlcolor  = WildStrawberry,
}

\usepackage[breaklinks=true,bookmarks=false]{hyperref}

\iccvfinalcopy 

\def\iccvPaperID{****} 

\ificcvfinal\fi

\usepackage[capitalize]{cleveref}
\crefname{section}{Sec.}{Secs.}
\Crefname{section}{Section}{Sections}
\Crefname{table}{Table}{Tables}
\crefname{table}{Tab.}{Tabs.}
\crefname{equation}{}{}

\newcommand{\methodname}{CLoVE\xspace}

\let\given\givenbase

\def\imagesize{2.45cm}

\newcommand\nomarkfootnote[1]{%
  \begingroup
  \renewcommand\thefootnote{}\footnote{#1}%
  \addtocounter{footnote}{-1}%
  \endgroup
}

\begin{document}

\title{Self-supervised Learning of Contextualized Local Visual Embeddings}

\author{Thalles Silva \\
\small{Institute of Computing} \\ \small{University of Campinas} \\
\small{Campinas-SP, Brazil} \\
{\tt\small thalles.silva@students.ic.unicamp.br}
\and
Helio Pedrini \\
\small{Institute of Computing} \\ 
\small{University of Campinas} \\
\small{Campinas-SP, Brazil} \\
{\tt\small helio@ic.unicamp.br}
\and
Ad\'in Ram\'irez Rivera \\
\small{Department of Informatics} \\
\small{University of Oslo} \\
\small{Oslo, Norway} \\
{\tt\small adinr@uio.no}
}

\maketitle
\nomarkfootnote{To appear in the 4th Visual Inductive Priors for Data-Efficient Deep Learning Workshop at ICCV 2023}
\ificcvfinal\thispagestyle{empty}\fi

\begin{abstract}
    We present Contextualized Local Visual Embeddings (\methodname), a self-supervised convolutional-based method that learns representations suited for dense prediction tasks.
    \methodname deviates from current methods and optimizes a single loss function that operates at the level of contextualized local embeddings learned from output feature maps of convolution neural network (CNN) encoders. 
    To learn contextualized embeddings, \methodname proposes a normalized mult-head self-attention layer that combines local features from different parts of an image based on similarity. 
    We extensively benchmark \methodname's pre-trained representations on multiple datasets.
    \methodname reaches state-of-the-art performance for CNN-based architectures in 4 dense prediction downstream tasks, including object detection, instance segmentation, keypoint detection, and dense pose estimation. Code: \url{https://github.com/sthalles/CLoVE}.
\end{abstract}

\section{Introduction}

Self-supervised learning (SSL) has become essential for learning downstream tasks. 
For tasks in which data annotation is pricey or even impossible to acquire, a round of self-supervised pre-training prior to learning the downstream task of interest can significantly enhance the system's final performance and reduce costs with data annotation.

In computer vision, one main advantage of SSL~\citep{oord2018representation,he2020momentum,chen2020simple,grill2020bootstrap} over generative models~\citep{goodfellow2020generative,vincent2008extracting,pathak2016context}, is the avoidance of reconstructing the input signal. 
Typically, generative models optimize a cost function in the pixel space, seeking to reconstruct the original input with high fidelity. 
Besides the high computing costs of operating in the pixel space, these methods assume that every pixel in the image matters equally. 
However, from the representation learning perspective, this property may not be necessary.  

Instead, the SSL approach of working at the embedding level allows SSL methods to learn representations that discard useless information. 
This strategy can be precious for learning downstream tasks since much of the details of an image may be useless for solving many downstream tasks. 
For instance, if the task of interest only requires a global signal, such as the class information, given a fixed-size feature vector, the encoder may be encouraged to discard low-level details, such as position, background, and orientation, in favor of features associated with the class information.

\begin{figure}[tb]
    \centering
    \resizebox{1.0\linewidth}{!}{%
    \begin{tikzpicture}

        \tikzmath{
            int \CELLSIZE;
            \CELLSIZE=\imagesize/4;
        }

        \begin{scope}
            \node[inner sep=0pt, draw=red, line width=0.5mm] (view1) {\includegraphics[width=\imagesize]{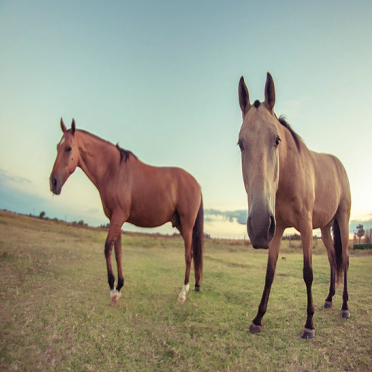}};
        
            \node [rectangle, draw, right=of view1, label=above:$\bar{F}$] (z1) {};
            \node [rectangle, draw, right=of z1, label=above:$\bar{F}$] (z2) {};
            
            \node[inner sep=0pt, draw=blue, line width=0.5mm, right=of z2] (view2) {\includegraphics[width=\imagesize]{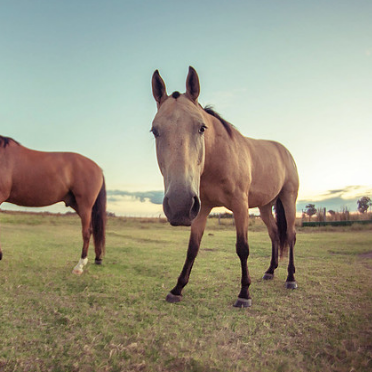}};
        
            \draw[dotted] (view1.south west) -- (z1.south west);
            \draw[dotted] (view1.south east) -- (z1.south east);
            \draw[dotted] (view1.north west) -- (z1.north west);
            \draw[dotted] (view1.north east) -- (z1.north east);
            \draw[dotted] (view2.south west) -- (z2.south west);
            \draw[dotted] (view2.south east) -- (z2.south east);
            \draw[dotted] (view2.north west) -- (z2.north west);
            \draw[dotted] (view2.north east) -- (z2.north east);
        
            \node [scale=0.2](midpoint) at ($(z1)!0.5!(z2)$) {};
            \draw[thick,->] (z1) -- (midpoint);
            \draw[thick,->] (z2) -- (midpoint);
        \end{scope}
            
        \begin{scope}
            \node[inner sep=0, below=of view1, yshift=25] (view1) {\includegraphics[width=\imagesize]{vi1.png}};
                 
            \node [rectangle, draw, right=of view1, label=above:$F_i$] (z1) {};
            \node [rectangle, draw, right=of z1, label=above:$F_j$] (z2) {};
            
            \node[inner sep=0pt, right=of z2] (view2) {\includegraphics[width=\imagesize]{vi2.png}};
        
            \matrix[draw, matrix of nodes, inner sep=0, left=of z1, nodes={draw, inner sep=0, minimum size=\CELLSIZE}, color=gray!10, opacity=0.5, nodes in empty cells, ampersand replacement=\&](box1) {
                \& \& \& \\
                \& \node(box1-2-2)[color=red, opacity=1.0]{}; \& \& \\
                \& \& \& \\
                \& \& \& \\
            };
        
            \matrix[draw, matrix of nodes, inner sep=0, right=of z2, nodes={draw, inner sep=0, minimum size=\CELLSIZE}, color=gray!10, opacity=0.5, nodes in empty cells, ampersand replacement=\&](box2) {
                 \& \& \& \\
                 \node(box2-2-1)[color=blue, opacity=1]{}; \& \& \& \\
                 \& \& \& \\
                 \& \& \& \\
            };        
        
            \draw[dotted] (box1-2-2.south west) -- (z1.south west);
            \draw[dotted] (box1-2-2.south east) -- (z1.south east);
            \draw[dotted] (box1-2-2.north west) -- (z1.north west);
            \draw[dotted] (box1-2-2.north east) -- (z1.north east);
            \draw[dotted] (box2-2-1.south west) -- (z2.south west);
            \draw[dotted] (box2-2-1.south east) -- (z2.south east);
            \draw[dotted] (box2-2-1.north west) -- (z2.north west);
            \draw[dotted] (box2-2-1.north east) -- (z2.north east);
            
            \node [scale=0.2](midpoint) at ($(z1)!0.5!(z2)$) {};
            \draw[thick,->] (z1) -- (midpoint);
            \draw[thick,->] (z2) -- (midpoint);
        \end{scope}
        
        \begin{scope}
            \node[inner sep=0pt, below=of view1, yshift=25] (view1) {\includegraphics[width=\imagesize]{vi1.png}};
            
            \node [rectangle, draw, right=of view1, label=above:$C_i$] (z1) {};
            \node [rectangle, draw, right=of z1, label=above:$F_j$] (z2) {};
            
            \node[inner sep=0pt, right=of z2] (view2) {\includegraphics[width=\imagesize]{vi2.png}};
            
            \matrix[draw, matrix of nodes, inner sep=0, left=of z1, nodes={draw, inner sep=0, minimum size=\CELLSIZE}, color=gray!10, opacity=0.5, nodes in empty cells, ampersand replacement=\&](box1) {
                 \node[fill=red!50]  {}; \& \node[fill=red!60] {}; \& \node[fill=red!70] {}; \& \node[fill=red!50] {}; \\
                 \node[fill=red!90] {};  \& \node(box1-2-2)[fill=red!100,color=red]{}; \& \node[fill=red!90] {}; \& \node[fill=red!80] {}; \\
                 \node[fill=red!60]  {}; \& \node[fill=red!80] {}; \& \node[fill=red!50] {}; \& \node[fill=red!40] {}; \\
                 \node[fill=red!5]   {}; \& \node[fill=red!2]  {}; \& \node[fill=red!0]  {}; \& \node[fill=red!0] {}; \\
            };
        
            \matrix[draw, matrix of nodes, inner sep=0, right=of z2, nodes={draw, inner sep=0, minimum size=\CELLSIZE}, color=gray!10, opacity=0.5, nodes in empty cells, ampersand replacement=\&](box2) {
                 \& \& \& \\
                 \node(box2-2-1)[color=blue, opacity=1]{}; \& \& \& \\
                 \& \& \& \\
                 \& \& \& \\
            };   
            
            \draw[dotted] (view1.south west) -- (z1.south west);
            \draw[dotted] (view1.south east) -- (z1.south east);
            \draw[dotted] (view1.north west) -- (z1.north west);
            \draw[dotted] (view1.north east) -- (z1.north east);
            \draw[dotted] (box2-2-1.south west) -- (z2.south west);
            \draw[dotted] (box2-2-1.south east) -- (z2.south east);
            \draw[dotted] (box2-2-1.north west) -- (z2.north west);
            \draw[dotted] (box2-2-1.north east) -- (z2.north east);
        
            \node [scale=0.2](midpoint) at ($(z1)!0.5!(z2)$) {};
            \draw[thick,->] (z1) -- (midpoint);
            \draw[thick,->] (z2) -- (midpoint);
                
        \end{scope}
    \end{tikzpicture}}
    \caption{SSL strategies to learn representations. Embedding similarity optimization over global representations (top), local representations (middle), and contextualized embeddings (bottom).}
    \label{fig:ssl-comp}
\end{figure}


Classic convolutional neural networks (CNNs) were primarily designed to address classification tasks. 
CNNs decimate the spatial dimensions of the input in favor of learning dense feature maps that are collapsed to a single global representation vector before going to a classifier layer. 
This engineering tendency encourages the convolutional encoder to discard fine-grained information from the input. 
In fact, that is why many segmentation models~\citep{chen2018encoder,long2015fully} attempt to reconstruct the input image, which can be viewed as learning the low-level details lost in the encoding process.

We argue that current SSL methods, based on CNN backbones, inherit the same architecture designs and suffer from similar problems. 
Collapsing the output feature maps of a CNN encoder into a global-level vector using an aggregation function, such as the average, encourages the encoder to discard low-level details crucial for solving dense prediction tasks, such as detection and segmentation.

Based on these assumptions, we conjecture that CNN-based SSL methods carry an engineering bias toward downstream tasks that do not require low-level information from the input. 
Such biases are also enforced by evaluation protocols that primarily assess the learned representation's classification power. 
For these reasons, state-of-the-art SSL methods perform much better in classification tasks than downstream tasks requiring dense predictions.

To close this gap, we propose an algorithmic approach that focuses on learning contextualized visual embeddings.
Contextualized embeddings combine local features of an image based on self-similarities. 
Instead of aggregating local feature maps into a global vector using an arithmetic average that attributes equal weights to each local feature, we bootstrap multiple prediction vectors (one for each local feature) based on learned weighted averages that capture contextualized information from similar regions of the input image, as illustrated in~\Cref{fig:ssl-comp}.
This way, we can bootstrap prediction vectors that aggregate multiple areas of an image view that share semantic meaning to predict local parts of a different view of the same image.
Our method, \textbf{C}ontextualized \textbf{Lo}cal \textbf{V}isual \textbf{E}mbeddings (\methodname), is designed to learn representations that preserve local information from the input by finding correlations among similar regions of a view to predict local parts of a different view. The motivation is to learn representations that excel at solving downstream dense prediction tasks.

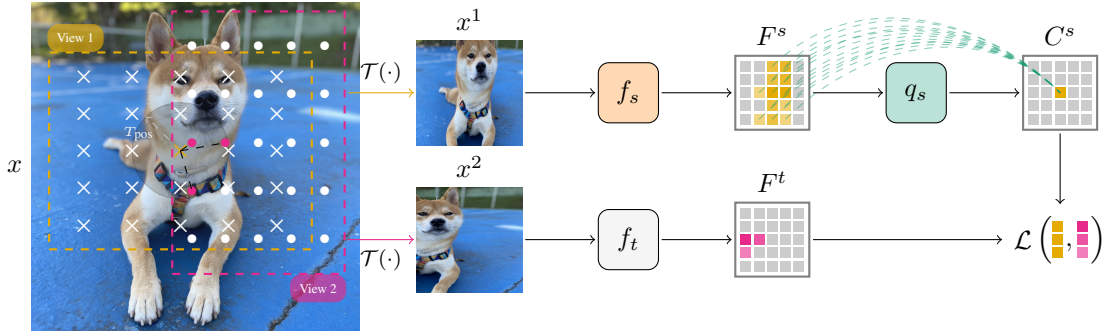
\begin{figure*}[tb]
  \centering
  \colorlet{v1c}{Dark2-F}
  \colorlet{v2c}{Dark2-D}
  \pgfdeclarelayer{bg}
  \pgfdeclarelayer{mg}
  \pgfsetlayers{bg,mg,main}
  \begin{tikzpicture}[
    module/.style={
        rectangle, 
        rounded corners, 
        minimum width=0.8cm, 
        minimum height=0.8cm, 
        text centered, 
        draw=black
    },
    arrow/.style={
        ->, 
        shorten >= 1pt,
        shorten <= 1pt,
    },
    matrix of feats/.style={
        matrix of nodes,
        nodes in empty cells,
        row sep=1pt, 
        column sep=1pt, 
        nodes={
            minimum width=4pt, 
            minimum height=4pt, 
            inner sep=0pt, 
            outer sep=0pt,
            fill=#1,
        },
        draw=gray,
        thick,
        inner sep=2pt,
    },
    matrix of feats/.default=black!20,
    set content/.style={
        execute at begin node once={#1},
    },
    matrix of symbs/.style n args={3}{
        execute at begin node={\nodesymb},
        execute at begin node once/.store in=\nodesymb,
        matrix of nodes,
        nodes in empty cells,
        column sep=#2,
        row sep=#3,
        nodes={
            text=white,
            font=\footnotesize,
            set content=#1,
            minimum width=1em,
            minimum height=1em,
            inner sep=0pt,
            outer sep=0pt,
        },
    },
    view/.style={
        rectangle,
        thick,
        draw,dashed, 
        color=#1,
    },
    view label/.style={
        fill=#1, 
        draw=#1,
        text=white, 
        rounded corners, 
        opacity=0.65, 
        text opacity=1, 
        font=\tiny,
        outer sep=0pt,
    },
  ]
  
    \newlength{\figsz}
    \setlength{\figsz}{0.25\textwidth}
    \begin{pgfonlayer}{bg}
    \node[inner sep=0pt, opacity=0.95, outer sep=0pt, label=left:$x$] (image) {\includegraphics[width=\figsz]{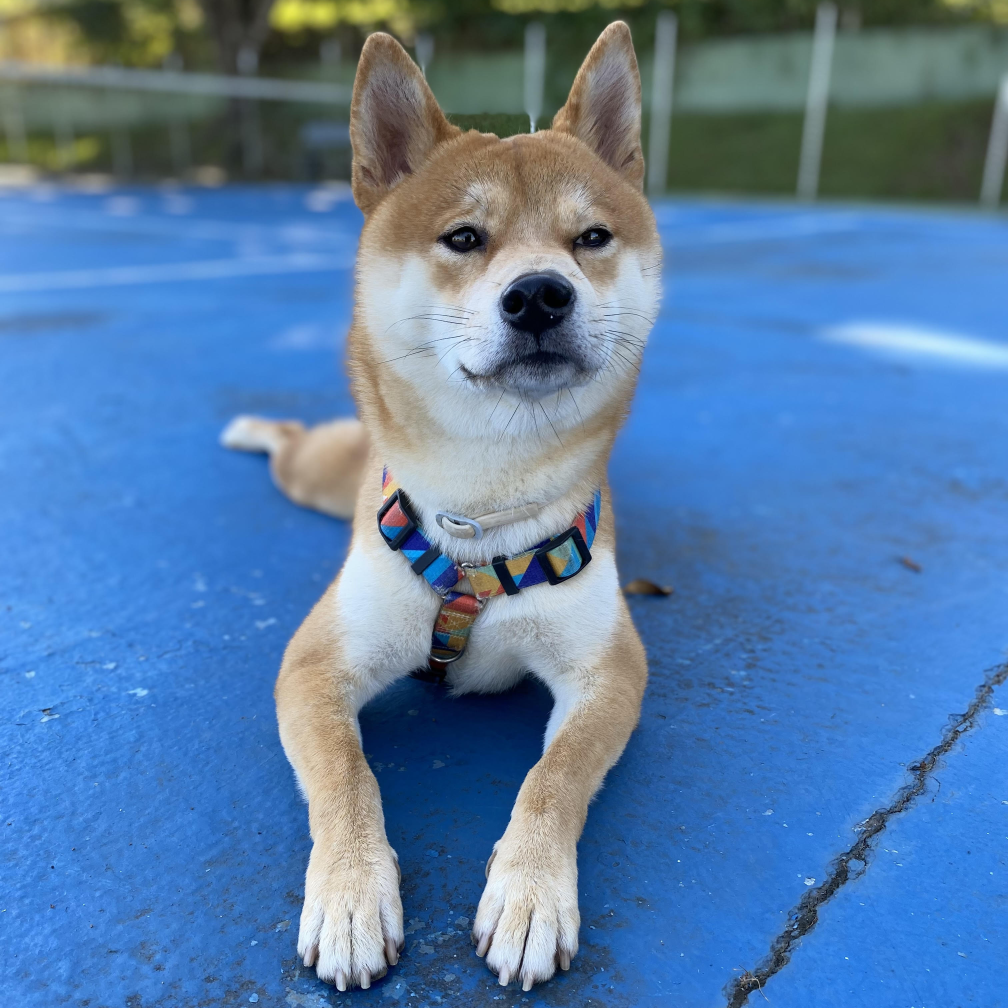}};
    \end{pgfonlayer}

    \node[view=v1c, minimum width=0.8\figsz, minimum height=0.6\figsz, anchor=north west] (box1) at ($(image.north west)+(0.05\figsz,-0.15\figsz)$) {};
    \node[anchor=south west, view label=v1c] at (box1.north west) {View 1};

    \node[matrix of symbs={$\boldsymbol{\times}$}{(0.8\figsz-5em)/6}{(0.6\figsz-5em)/6}] (v1-sym) at (box1) {
        &&&&\\
        &&&&\\
        &&|[v1c]|&&\\
        &&&&\\
        &&&&\\
    };

    \node[view=v2c, minimum width=0.525\figsz, minimum height=0.8\figsz, anchor=north west] (box2) at ($(image.north west)+(0.425\figsz,-0.025\figsz)$) {};
    \node[anchor=north east, view label=v2c] at (box2.south east) {View 2};
    
    \node[matrix of symbs={$\bullet$}{(0.525\figsz-5em)/6}{(0.8\figsz-5em)/6}] (v2-sym) at (box2) {
        &&&&\\
        &&&&\\
        |[v2c]|&|[v2c]|&&&\\
        |[v2c]|&&&&\\
        &&&&\\
    };
        
    \node[inner sep=0pt, above right=-0.5cm and 0.75cm of image, anchor=north west, label=above:$x^1$] (view1) {\includegraphics[width=.08\textwidth]{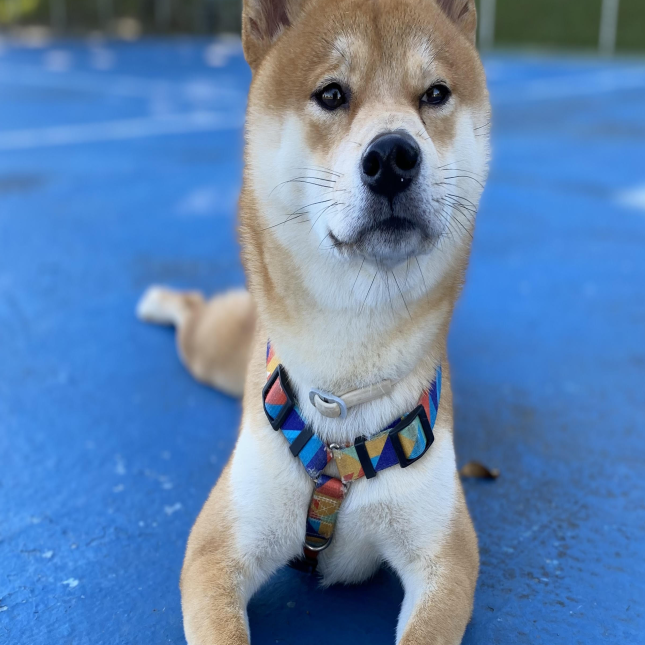}};
    
    \node[inner sep=0pt, below right=-0.5cm and 0.75cm of image, anchor=south west, label=above:$x^2$] (view2){\includegraphics[width=.08\textwidth]{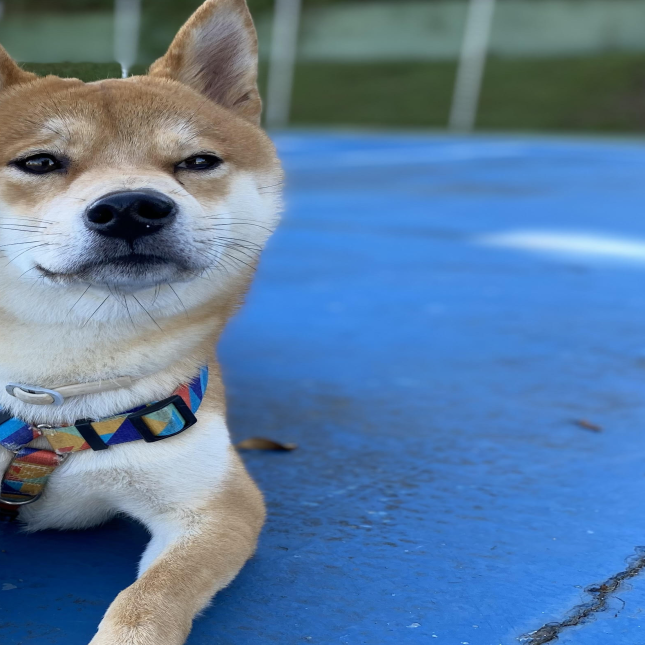}};

    \node [module, fill=orange!30, right=of view1] (enc1) {$f_s$};
    
    \node [module, fill=gray!8, right=of view2] (enc2) {$f_t$};
    
    \node [matrix of feats, right=of enc1, label=above:$F^s$] (proj_q) {
        &&|[v1c!75]|&|[v1c!85]|&\\
        &&|[v1c!75]|&|[v1c!85]|&\\
        &|[v1c!50]|&|[v1c]|&|[v1c]|&\\
        &&|[v1c!75]|&|[v1c!55]|&\\
        &|[v1c!50]|&|[v1c!75]|&|[v1c!50]|&\\
    };
    
    \node [matrix of feats, right=of enc2, label=above:$F^t$] (proj_k) {
        &&&&\\
        &&&&\\
        |[v2c]|&|[v2c!80]|&&&\\
        |[v2c!60]|&&&&\\
        &&&&\\
    };
    
    \node [module, fill=Dark2-A!30, right=of proj_q] (base-attn) {$q_s$};

    \node [matrix of feats, right=of base-attn, label={[align=center]$C^s$}] (ctx_embed) {
        &&&&\\
        &&&&\\
        &&|[v1c]|&&\\
        &&&&\\
        &&&&\\
    };
    
            
    \node [] at (ctx_embed|-proj_k) (ctx2loc) {$\mathop{\mathcal{L}} \Big( \hspace{.5em}, \hspace{.5em} \Big)$};
    \node [matrix of feats=v1c, draw=none, left=-3.5pt of ctx2loc.center] (e1) {
        \\ 
        \\ 
        \\ 
    };
    \node [matrix of feats, draw=none, right=4pt of ctx2loc.center] (e2) {
        |[v2c]|\\ 
        |[v2c!80]|\\ 
        |[v2c!60]|\\ 
    };

    \draw[arrow, draw=v1c] (box2.east |- view1.west) -- node[above, font=\footnotesize] {$\mathcal{T}(\cdot)$} (view1.west);
    \draw[arrow, draw=v2c] (box2.east |- view2.west) -- node[below, font=\footnotesize] {$\mathcal{T}(\cdot)$} (view2.west);

    \draw[arrow] (view1.east) -- (enc1.west);
    \draw[arrow] (view2.east) -- (enc2.west);

    \draw[arrow] (enc1.east) -- (proj_q.west);
    \draw[arrow] (enc2.east) -- (proj_k.west);
    
    \draw[arrow] (proj_q.east) -- (base-attn.west);
    \draw[arrow] (base-attn.east) -- (ctx_embed.west);
    \draw[arrow] (proj_k.east) -- (ctx2loc.west);
    \draw[arrow] (ctx_embed.south) -- (ctx2loc.north);
    \begin{pgfonlayer}{mg}
        \draw[fill=black!75, opacity=0.25] let \p1 = ($(v1-sym-3-3.center)-(v2-sym-3-2.center)$) in (v1-sym-3-3) circle ({veclen(\x1,\y1)+1pt});    
        \draw[dashed, black!25, ->] let \p1 = ($(v1-sym-3-3.center)-(v2-sym-3-2.center)$) in (v1-sym-3-3.center) -- ++(135:{veclen(\x1,\y1)+1pt}) node[midway, left, font=\tiny, text=black!5] {$T_{\text{pos}}$};
    \end{pgfonlayer}

    \draw[dashed] (v1-sym-3-3.center) -- (v2-sym-3-1.center);
    \draw[dashed] (v1-sym-3-3.center) -- (v2-sym-3-2.center);
    \draw[dashed] (v1-sym-3-3.center) -- (v2-sym-4-1.center);

    \foreach \p in {1-3,1-4,2-3,2-4,3-2,3-3,3-4,4-3,4-4,5-2,5-3,5-4}{
        \draw[dashed, Dark2-A, opacity=.5] (proj_q-\p.center) to [in=135] (ctx_embed-3-3.center);
    }
    
    \end{tikzpicture}
    \caption{Views $x^1$ and $x^2$ are fed to student and teacher encoders $f_s$ and $f_t$, to extract local feature maps $F^s$ and $F^t$, respectively. The predictor $q_s$ takes the local features $F^s$ and outputs contextualized embeddings $C^s$ by combining local features based on self-similarities. We define a grid of points proportional to the output feature map in each view. Points in one view are paired with points in the other based on distance in the ambient space. Selected points are mapped to the feature space and used to match embeddings in $C^s$ with targets in $F^s$.}
    \label{fig:overview}
\end{figure*}

Traditional SSL methods primarily optimize global representations of different views on an image~\citep{he2020momentum,bardes2021vicreg,grill2020bootstrap}. 
When training CNN backbones, the output feature map is collapsed using an average function and treated as a global image representation. 
Conversely, current SSL methods designed for dense prediction representation learning~\citep{o2020unsupervised,xiao2021region,wang2021dense} either optimize for local features or combine local and global objectives. 
In contrast, \methodname does not optimize directly for local or global representations. 
Instead, it poses the representation learning problem at the level of contextualized local embeddings. 
We propose an objective function that predicts a target representation from a local part of a view using a combination of correlated local embeddings from another view. \Cref{fig:overview} illustrates our architecture.

Our contributions are twofold. Firstly, we introduce a novel method that does not optimize for local or global embeddings. 
Secondly, we propose a variation of the self-attention algorithm and integrate it into CNN architectures. 
Our method learns representations that effectively retain local information from the input and capture long-range dependencies from representations that share semantic meaning. 
This integration empowers our approach to excel in dense prediction downstream tasks, where fine-grained details play a vital role in achieving high performance and accuracy. 
Our method is extensively evaluated and proves its effectiveness in downstream tasks, including object and keypoint detection, segmentation, and pose estimation.

\section{Related work}

Recent SSL methods follow a similar framework composed of the following building blocks: (1)~a joint-embedding architecture, (2)~a pretext task, and (3)~a similarity-based loss function. The joint-embedding architecture may be pure siamese~\citep{bromley1993signature} or follow a teacher-student~\citep{chen2020big} architecture with a separate momentum encoder that usually does not receive gradients. Among many proposed pretext tasks, one that stands out is instance discrimination~\citep{wu2018unsupervised,alexey2015discriminative}. 
For instance discrimination, we task a deep neural network to find a pair of representations from different views of the same image among a set of negative pairs where the representation from the anchor image is paired with representations from random images. Lastly, the similarity loss function may be contrastive~\citep{he2020momentum,chen2020simple,tian2020makes}, in which InfoNCE~\citep{oord2018representation} is a popular choice, or non-contrastive~\citep{grill2020bootstrap,chen2021exploring}. 

SSL methods differ in how they optimize the embedding space. While a group of methods directly optimize the representations using a similarity loss function~\citep{chen2020simple,he2020momentum,zbontar2021barlow}, others discretize the embedding space by learning prototypes~\citep{asano2019self,caron2018deep,silva2022representation,caron2020unsupervised}. Despite differences, these methods are designed to learn global representations from the input image. When the feature extractor is represented as a CNN, the feature map from the last convolutional layer is collapsed into a single vector through a global average pooling operation. If a Transformer~\citep{dosovitskiy2020image} backbone is used, the class-token representation is optimized as a global feature vector~\citep{caron2021emerging,chen2021empirical}. These methods generally learn powerful, invariant representations for classification problems but do not perform as well when the downstream task requires localization and low-level details.

Recently, we have witnessed the emergence of methods designed for dense prediction tasks~\citep{o2020unsupervised, wang2021dense, bardes2022vicregl, xie2021propagate}. Generally, these methods take one of two approaches to learn representations (1)~they pose the learning problem at the level of local embeddings~\citep{o2020unsupervised}, or (2)~they optimize for global and local embeddings jointly~\citep{bardes2022vicregl,wang2021dense,xie2021propagate,xiao2021region}.
Most methods fall into the second category, where two loss functions are minimized, one that operates on representations from the full view and another on representations from local parts of the image. The two loss functions are linearly combined to a final objective and jointly optimized. Some evidence suggests a trade-off between global and local feature learning for SSL~\citep{bardes2022vicregl,wang2021dense}, which might explain the popular algorithmic design. We can view this approach as an extension of current SSL methods, allowing them to trade off global and local characteristics in their learning features.

Among methods that pose the learning problem at the local feature level, the approach proposed by \citet{o2020unsupervised} stands out. The method learns dense (pixel-level) representations by exploring contrastive learning over local features that map to the same pixel across different views of the same image. The architecture learns local features by reconstructing the feature maps using a decoder model and applies contrastive learning at a higher level of feature reconstruction.

Among methods that combine global and local objectives, recent work~\citep{wang2021dense,wang2022exploring,xiao2021region} used the InfoNCE loss to learn global and local representations and can be viewed as extensions of MoCo~\citep{he2020momentum}. 
\citet{wang2021dense} proposed a loss function that performs contrastive learning at the level of local features. 
To match local features across different views, they use a cosine similarity function where a local feature from one view takes the most similar local feature from the other view as its target. 
Similarly, \citet{xiao2021region} proposed a region-level contrastive loss that relies on intersected regions between the two views of an image. 
Over intermediate layers of a convolutional encoder, the overlapping areas (feature maps) are processed by a fixed-sized window and fed to a Precise RoI Pooling~\citep{jiang2018acquisition} layer, creating a feature vector from the region. 
In both cases, the local loss is implemented using the InfoNCE loss and jointly optimized with the global MoCo-style objective.
 
\citet{xie2021propagate} proposed a non-contrastive local objective that can be viewed as an extension to the BYOL~\cite{grill2020bootstrap} loss. 
They proposed the Pixel-to-Propagation module. A form of attention layer that creates contextualized local embeddings by combining local features in a vicinity. 
Lastly, \citet{bardes2022vicregl} extended the VicReg~\citep{bardes2021vicreg} method and applied the Variance-Invariance-Covariance Regularization (VICReg) loss to learn global and local features.


\noindent\textbf{Contrast to previous approaches.}
Our method differs from contemporary work in essential aspects. One of the main differences between \methodname and existing approaches is the departure of jointly optimizing global and local objectives, thus avoiding the global/local feature learning trade-off. Instead, we learn multi-head self-attention layers that can bootstrap contextualized local embeddings that serve as predictions to target local features.

\methodname may be regarded as similar to PixPro~\citep{xie2021propagate}. 
However, there are essential differences between the two approaches. 
\methodname combines multi-head self-attention layers, usually employed in transformers, to convolutional architectures in a contextualized local feature learning framework. 
On the other hand, the Pixel-to-Propagation module~\citep{xie2021propagate} differs from \methodname in important aspects. 
Namely, (1)~it does not learn multiple heads, (2)~it does not learn transformation matrices for query, key, and value tensors, and (3)~it does not normalize the result attention scores. 
Moreover,~\citet{xie2021propagate} combined a loss function at the local embedding with the standard BYOL global objective in a non-contrastive manner. 
Conversely, \methodname does not work directly with global or local objectives and employs a ranking margin loss.

Unlike previous work~\citep{o2020unsupervised}, our architecture works directly at the feature map level and does not attempt to reconstruct local features. 
In contrast to~\citet{wang2021dense}, our strategy avoids the noisy process of choosing the most similar local embedding as the target. 
Instead, we match representations from which their center pixels lie within a vicinity in the pixel space.

\section{Learning contextualized local representations}

We strive to learn visual features that retain fine-grained details from the input and therefore are suited for dense prediction tasks. 
Unlike other methods, \methodname does not optimize a global or a local loss function (or their combination). 
Instead, the learning problem is posed at the contextualized embeddings level, learned from feature maps of CNN encoders. 
In this framework, we use local features as target representations, and to predict such targets, we learn vectors that combine local features in a vicinity based on learned self-similarities. 
In essence, contextualized embeddings are a mixture of local, semantically similar features from different parts of a view. 
Local features are combined into a single prediction based on their similarity to the anchor local feature. 
Intuitively, this strategy allows learning richer prediction vectors that encode many similar parts of an image view to predict a localized portion of another view.

\subsection{Preliminares}

Given an image $x \in \mathbb{R}^{3 \times H \times W}$ with no supervision, we create views $x^1 = \mathcal{T}(x)$ and $x^2 = \mathcal{T}(x)$, where $\mathcal{T}(\cdot)$ is a stochastic function that applies a set of random geometric and intensity transformations to $x$. Such transformations include random flips, color distortions, and cropping. 
In practice, we can work with many views, but for simplicity, we constrain the number of views to $N_v=2$.

Each view is independently forwarded through a student encoder $f_s$ and a teacher encoder $f_t$. 
The encoders are composed of a feature extractor, \eg, a CNN encoder, and a projection head represented as a multi-layer perceptron (MLP). 
Following previous work~\cite{he2020momentum,grill2020bootstrap}, the teacher encoder $f_t$ does not receive gradient updates. 
Instead, the weights $\theta_t$ are updated using a moving average of the weights $\theta_s$, such as $\theta_t = \alpha \theta_t + (1-\alpha) \theta_s$, where $\alpha$ is the weight.

For each view, we obtain a tensor of projected local feature maps $F = f(x^v)$, for $v \in [0,1]$.
These local features correspond to the output feature map of an intermediate layer of the CNN feature extractor, projected to a lower dimensional space, and have a general shape of $F \in \mathbb{R}^{N \times D \times F_h \times F_w}$, where $N$ is the batch size, $D$ is the feature dimensionality, and $F_h$ and $F_w$ are the spatial dimensions of the feature map. 

We can view the projected local features in $F$ as a sequence of embeddings, $F \in \mathbb{R}^{N \times D \times L}$, where $L$ is the sequence length $L= F_h \times F_w$. 
Traditional SSL methods take the feature maps from the CNN feature extractor (prior to projection) and collapse them using a global average operation to obtain a global representation.
The global feature is fed to a projection head and then to a similarity-based loss function,  as illustrated in~\Cref{fig:ssl-comp} (top). 
On the other hand, local SSL methods either maximize agreement between local embeddings or combine local and global objectives~\cite{wang2021dense,xie2021propagate,bardes2022vicregl}. 
In a different direction, \methodname learns contextualized representations through self-attention layers operating on local embeddings of a view. 

Next, we detail how we extract dense self-supervision from image views and our contextualized loss function.

\subsection{Pixel-to-representation neighborhood matching}
\label{sec:self-supervised-local-feature-matching}

To learn representations that retain low-level features, we need targets that contain such properties. 
In other words, we must bootstrap dense self-supervised signals to use as targets in our loss function.
One way is to track pixels' locations as we create views $x^1$ and $x^2$.
If two views share an intersected area, the pixels in this region represent the same part in the original image. 
However, scaling and resizing may push these pixels to random locations during the view's creation.  
Instead of matching exact pixels across views, we can look for pixels' neighbors. 
This strategy explores the pixel spatial locality inductive bias in which nearby pixels represent similar contexts and, hence, should have similar representations.
Once we match pixels across views based on neighborhood distances, we can map the pixels' locations to the feature space to index local features in the loss function. 

We define $I^1$ and $I^2$ as lists of 2D points in the pixel space. 
Points in $I^1$ are defined over the first view, and points in $I^2$ over the second. 
For each point $I^1_i$ in the first view, we look for pixel correspondences in the second view by extracting nearby points in $I^2$ that lie within a similarity region. 
Accordingly, we define $M$ as the set of all pairs $(I^1_i, I^2_j)$ such that the euclidian distance between points $I^1_i$ and $I^2_j$ is smaller than a threshold $T_{\text{pos}}$, such as
\begin{equation}
M = \left\{ \left(I^1_i, I^2_j \right) \given d \left(I^1_i, I^2_j \right) < T_{\text{pos}} \right\},
\end{equation}
where $d(a,b) = \sqrt{\sum_{i=0}^2 \left ( a_i, b_i \right )^2}$.

Next, we map the points in $M$ from the pixel space to the feature space. 
Each point in $M$ is mapped to its respective local embedding in the feature map of the CNN encoder. 
Therefore, the pair of points in $M$ now represent a pair of indices matching features from view \num{1} to view \num{2}.
This process is depicted in~\Cref{fig:overview}.

The Pixel-to-Neighborhood matching strategy will pair at most $p = |F|$ points for each local embedding, where $F$ represents the projected feature map from the CNN encoder. 
For a ResNet-50 encoder, we define \num{49} points in a grid structure that are mapped to each of the $7 \times 7$ local features in $F$, as described in~\Cref{sec:impl-details}. 

One advantage of this matching algorithm is that we do not need to force views to share an intersected region. Local representations from different views that do not intersect can still be paired if they are close enough in the pixel space.
Moreover, the choice of $T_{\text{pos}}$ matters since it controls the average number of target local representations.
Intuitively, if $T_{\text{pos}}$ is too high, a pixel $I^1_i$ might consider all pixels in $I^2$ as neighbors. 
As a result, it invalidates the spatial locality inductive bias present in natural images.
On the other hand, if $T_{\text{pos}}$ is too low, it limits the target space as the spatial locality bias is not explored to its fullest, as described in~\Cref{sec:positive-theshold}. 

\subsection{Predicting local embeddings with contextualized vectors}
\label{sec:context-to-local}

At this point, we could match local features across different views on an image using the feature indices in $M$.
However, this learning objective would fail to learn long-range dependencies.
Intuitively, if an object occupies a large portion of an image, we want to maximize the agreement between all semantically meaningful parts of the object or region and its local target embedding.
To accomplish this strategy, each local feature of the first view can interact with its neighboring local features to learn similarity patterns.
This way, local features exhibiting substantial similarity are combined into a single contextualized vector and used to predict the local target embedding from another view. 

To learn contextualized embeddings, we propose a predictor head $q_s$ that receives the output feature map $F^s$ from the student and apply a Normalized Multi-Head Self-Attention (NMHSA) layer to obtain $C^s = q_s(F^s)$, where $q_s(F^s) = \textup{NMHSA}(F^s)$.
We use the matching feature indices in $M$ to select contextualized predictions and target local features from $C^s$ and $F^t$, respectively. Then, we maximize agreement between contextualized and local embeddings by minimizing the margin ranking loss defined as,
\begin{equation}
\label{eq:ctx-to-local-loss}
\mathcal{L} = \smashoperator{\sum_{\left( i, j \right) \in M}} \max \left(0, - \lambda \sigma \left(C^s_i, F^t_j \right) + \sigma \left(C^s_i, F^t_{\text{neg}} \right) + \mu \right),
\end{equation}
\noindent where $\mu$ is the margin, $\sigma(a, b) = \frac{x y}{\left \| x \right \|_2 \left \| y \right \|_2}$ is the cosine similarity function and $\left \| \cdot  \right \|_2 $ is the $\ell_2$ norm.

For each pair of matching features indexed by $(I^1_i, I^2_j)$, we maximize agreement between contextualized representations from one view and local embeddings from the other.

To bootstrap the negative representation $F_{\text{neg}}$, we follow a similar strategy proposed by \citet{wang2021solving}. 
We compute the cosine similarity between the contextualized predictions $C^s$ and all local representations from the opposing view $F^t$. 
Then, we select the top-$k$ most offending local representations (higher similarities scores) from $F^t$, discard the most similar one, and take the average of the resulting vectors. 
Intuitively, we discard the most offending local feature from $F^t$ because it could represent a false negative. 
This selection strategy can be viewed as finding a \textit{negative} region (within the image) that is not correlated with the contextualized predictor.
The size of the negative region is controlled by $k$ and set as $k=10$.
We show in \Cref{sec:negative-sampling-strategies} that choosing negatives within the image is most beneficial to the learned representation as selecting negatives across different images. 

\subsection{The normalized attention head}

We can view the self-attention mechanism as combining similar local areas of a view.
Intuitively, to successfully predict the local region of the second view, the self-attention must combine the local features of the first view in a way that similar content has a strong contribution and dissimilar content has a weak contribution to the contextualized embedding. 

In practice, we learn $8$ self-attention heads, where $\textup{head}[i] = \textup{Attention}(F^sW^q, F^sW^k, F^sW^v)$ and $\textup{Attention(Q,K,V)} = \textup{softmax}\left ( \frac{\sigma(Q,K^T)}{\tau} \right )V$.
We show in \Cref{sec:normalized-mhsa} that, in practice, normalizing queries and keys before computing the attention scores improves the final downstream tasks' performance.

From an intuitive perspective, by matching contextualized representations with local embeddings (based on pixel spatial locality), the network learns to (1)~attend to similar regions in the input and (2)~disregard local embeddings representing different contexts in the same view. 
This process optimizes multiple prediction subtasks, \ie, for each local feature $F^s_i$, there is a contextualized representation $C^s_i$. 
As a result, the learned representations retain fine-grain details from the input.

\section{Main experiments}

To assess how well \methodname's~pre-trained representations transfer to dense prediction tasks, we fine-tuned detection and segmentation models, using \texttt{Detectron2}~\cite{wu2019detectron2}, on Pascal VOC07, COCO, LVIS, and Cityscapes datasets.
For the competing methods, we used the officially released model checkpoints and reported performance metrics from their papers if the same evaluation protocol. Otherwise, we ran experiments in-house.
We pre-trained \methodname on the ImageNet-1M dataset for \num{200} and \num{400} epochs and compare its performance against state-of-the-art SSL methods on various downstream tasks such as object detection, instance segmentation, keypoint detection, and dense pose estimation.
The experiments report average performance across \num{5} independent runs. 
We highlight the top-1 performing methods in \textbf{bold} and top-2 \uline{underlined}. 

\paragraph{COCO detection and instance segmentation.}
Tables~\ref{tab:coco-eval-c4} and~\ref{tab:coco-eval-fpn} compare \methodname's performance using the R50-C4 and R50-FPN backbones against other methods. 
For the two backbones, \methodname achieved top-1 performance across both tasks.
Additionally, \methodname reached top-2 performance in \num{5} out of the \num{6} for R50-C4 and 4 out of 6 for R50-FPN in low-resource training settings. 

\begingroup
\setlength{\tabcolsep}{4.6pt}
\begin{table}[tb]
    \centering
    \sisetup{
        table-format=2.1,
        round-mode = places,
        round-precision=1,
        detect-all,
    }
    \caption{Obj.\ detection and segmentation on COCO (R50-C4).}
    \label{tab:coco-eval-c4}
    \footnotesize
    \begin{tabular}{lcSSSSSS}
        \toprule
        Method & ep & AP$^{\text{bb}}$ & $\text{AP}^{\text{bb}}_{50}$ & $\text{AP}^{\text{bb}}_{75}$ & AP$^{\text{mb}}$ & $\text{AP}^{\text{mb}}_{50}$ & $\text{AP}^{\text{mb}}_{75}$ \\
        \midrule
        Supervised & 100 & 38.2 & 58.2 & 41.2 & 33.3 & 54.7 & 35.2 \\
        Rand init & {--} & 26.4 & 44 & 27.8 & 29.3 & 46.9 & 30.8 \\
        \midrule
        ReSim~\citep{xiao2021region} & 200 & 39.7 & 59 & 43 & 34.6 & 55.9 & 37.1\\
        InsCon~\citep{yang2022inscon} & 200 & 40.3 & \uline{60.0} & 43.5 & 35.1 & 56.7 & 37.6 \\
        PixPro~\citep{xie2021propagate} & 400 & 40.5 & 59.8 & 44 & \uline{35.4} & \uline{56.9} & 37.7 \\
        DetCo~\citep{xie2021detco} & 200 & 39.8 & 59.7 & 43 & 34.7 & 56.3 & 36.7 \\
        SlotCon~\citep{wen2022self} & 200 & 39.9 & 59.8 & 43.0 & 34.9 & 56.5 & 37.3 \\
        \midrule
        \methodname & 200 & \uline{40.6} & \uline{60.0} & \uline{44.1} & \uline{35.4} & 56.8 & \uline{37.8} \\
                    & 400 & \bfseries 41.0 & \bfseries 60.3 & \bfseries 44.2 & \bfseries 35.5 & \bfseries 57.2 & \bfseries 38.1\\
        \bottomrule
    \end{tabular}
\end{table}
\endgroup

\begingroup
\setlength{\tabcolsep}{4.6pt}
\begin{table}[tb]
    \centering
    \sisetup{
        table-format=2.1,
        round-mode = places,
        round-precision=1,
        detect-all,
    }
    \label{tab:coco-eval-fpn}
    \caption{Obj.\ detection and segmentation on COCO (R50-FPN).}
    \footnotesize
    \begin{tabular}{lcSSSSSS}
        \toprule
        Method & ep & AP$^{\text{bb}}$ & $\text{AP}^{\text{bb}}_{50}$ & $\text{AP}^{\text{bb}}_{75}$ & AP$^{\text{mb}}$ & $\text{AP}^{\text{mb}}_{50}$ & $\text{AP}^{\text{mb}}_{75}$ \\
        \midrule
        Supervised & 100 & 38.9 & 59.6 & 42.7 & 35.4 & 56.5 & 38.1\\
        Rand init  & {--} & 32.8 & 51 & 35.3 & 28.5 & 46.8 & 30.4\\
        \midrule
        DenseCL~\citep{wang2021dense} & 200 & 39.4 & 59.9 & 42.7 & 35.6 & 56.7 & 38.2\\
        ReSim~\citep{xiao2021region} & 200 & 39.3 & 59.7 & 43.1 & 35.7 & 56.7 & 38.1\\
        PixPro~\citep{xie2021propagate} & 400 & 39.8 & 59.5 & 43.7 & 36.1 & 56.5 & 38.9\\
        SetSim~\citep{wang2022exploring} & 200 & 40.2 & \uline{60.7} & 43.9 & 36.4 & \uline{57.7} & 39\\
        VICRegL~\citep{bardes2022vicregl} & 300 & 37.3 & 57.6 & 40.7 & 34.1 & 54.7 & 36.5\\
        \midrule
        \methodname & 200 & \uline{40.8} & 60.5 & \uline{45.0} & \uline{36.8} & 57.6 & \uline{39.8}\\
                    & 400 & \bfseries 41.2 & \bfseries 61.1 & \bfseries 45 & \bfseries 37.1 & \bfseries 58.1 & \bfseries 40.1\\
        \bottomrule
    \end{tabular}
    
\end{table}
\endgroup


\paragraph{Cityscapes instance segmentation.}
In \Cref{tab:cityscapes-eval}, 
\methodname achieves an average improvement of \textbf{+1.4} AP over PixPro~\citep{xie2021propagate}, and \textbf{+10.7} AP over the supervised baseline.

\begingroup
\begin{table}[tb]
  \centering
  \sisetup{
    table-format=2.1,
    round-mode = places,
    round-precision=1,
    detect-all,
  }
  \caption{Instance segmentation on Cityscapes (R50-FPN).}
  \label{tab:cityscapes-eval}
  \footnotesize
  \begin{tabular}{lcSS}
    \toprule
    Method & ep & AP & $\textup{AP}_{50}$ \\
    \midrule
    Supervised & 100 & 26.5 & 52.9 \\
    Rand init  & {--} & 19.9 & 40.7 \\
    \midrule
    DenseCL~\citep{wang2021dense} & 200 & 33.1 & 61.7 \\
    PixPro~\citep{xie2021propagate} & 400 & \uline{35.8} & 63.7 \\
    VICRegL~\citep{bardes2022vicregl} & 300 & 29.8 & 58.5 \\
    SlotCon~\citep{wen2022selfsupervised} & 200 & 35.2 & 63.8 \\    
    \midrule
    \methodname & 200 & 35.7 & \uline{64.1} \\
                & 400 & \bfseries 37.2 & \bfseries 65.3 \\
    \bottomrule
  \end{tabular}
\end{table}
\endgroup

\paragraph{LVIS object detection and instance segmentation.}

LVIS is a dataset for long-tail object recognition. 
It contains more than \num{1200} classes and more than \num{2}M high-quality instance segmentation masks. 
In \Cref{tab:lvis-eval},~\methodname \num{200} epoch model performs similarly to PixPro. 
The \num{400} epoch model beats competitors by a small margin and improves upon the supervised baseline by \textbf{+4} points in all metrics.

\begingroup
\setlength{\tabcolsep}{4.2pt}
\begin{table}[tb]
    \centering
    \sisetup{
        table-format=2.1,
        round-mode = places,
        round-precision=1,
        detect-all,
    }
    \caption{Obj.\ detection and segmentation on LVIS (R50-FPN).}
    \label{tab:lvis-eval}
    \footnotesize
    \begin{tabular}{lcSSSSSS}
        \toprule
            Method & ep & AP$^{\textup{bb}}$ & $\textup{AP}^{\textup{bb}}_{50}$ & $\textup{AP}^{\textup{bb}}_{75}$ & AP$^{\textup{mb}}$ & $\textup{AP}^{\textup{mb}}_{50}$ & $\textup{AP}^{\textup{mb}}_{75}$ \\
        \midrule
        Supervised & 100 & 20.2 & 33.4 & 21.4 & 19.6 & 31.2 & 20.8\\
        Rand init & {--}  & 12.4 & 21.8 & 12.5 & 12.1 & 20.2 & 12.5 \\
        \midrule
        DenseCL~\citep{wang2021dense} & 200 & 20.4 & 33.5 & 21.4 & 19.9 & 31.5 & 20.9 \\
        PixPro~\citep{xie2021propagate} & 400 & \uline{23.8} & \uline{38.2} & \uline{25.2} & \uline{23.3} & \uline{36.1} & 24.7\\
        SlotCon~\citep{wen2022self} & 200 & 23.2 & 37.6 & 24.3 & 22.9 & 35.6 & 24.3\\
        VICRegL~\citep{bardes2022vicregl} & 200 & 7 & 13.4 & 6.4 & 7.4 & 12.7 & 7.3\\
        \midrule
        \methodname & 200 & 23.6 & 37.7 & \uline{25.2} & \uline{23.3} & 35.9 & \uline{24.8} \\
                    & 400 & \bfseries 24.3 & \bfseries 38.8 & \bfseries 25.8 & \bfseries 23.9 & \bfseries 36.7 & \bfseries 25.3\\
        \bottomrule
    \end{tabular}
\end{table}
\endgroup

\paragraph{COCO keypoint detection.}

\begin{figure*}[tb]
    \centering
   \includegraphics[width=0.9\linewidth]{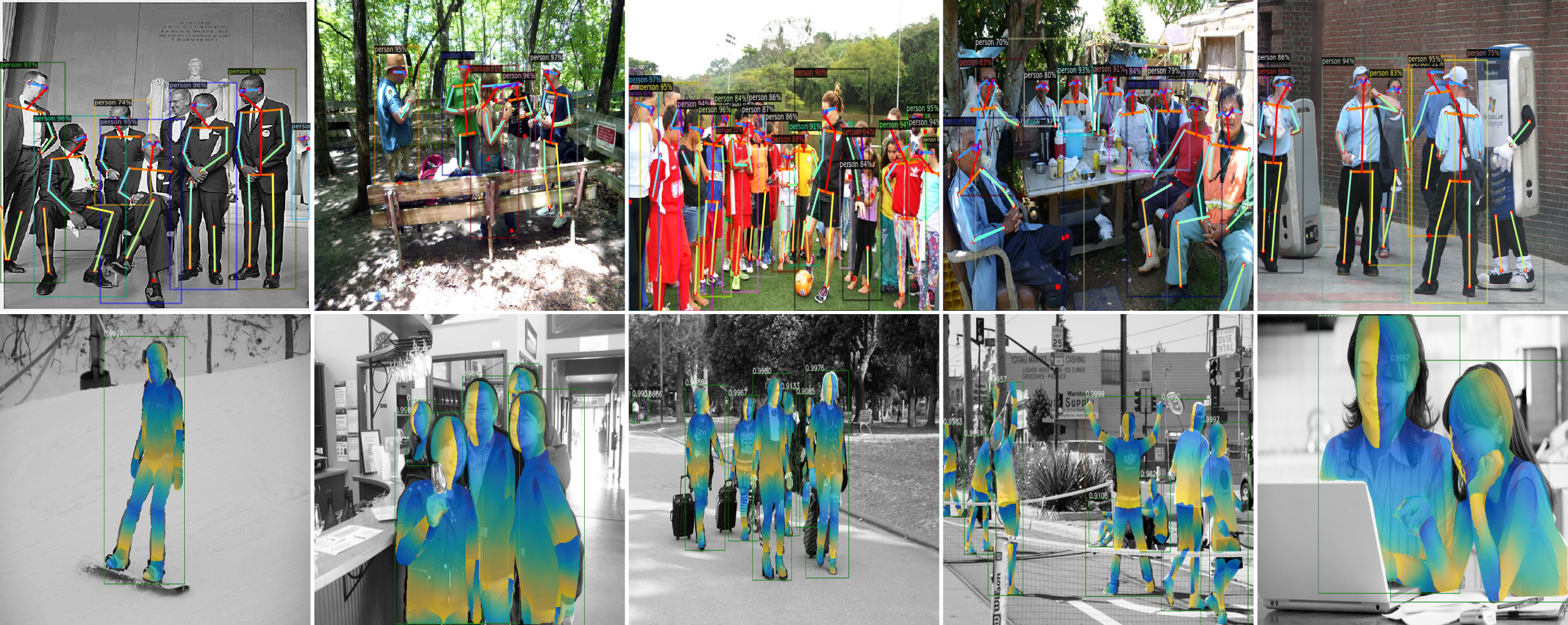}
   \caption{Qualitative results for keypoint detection (top row) and dense pose estimation (bottom row).}
   \label{fig:keypoints-dense-pose}
\end{figure*}

In \Cref{tab:keypoint-coco-eval}, \methodname performs comparably to other SSL methods and surpasses the supervised baseline by \textbf{+1.7} average AP.
For keypoint detection, we noticed that the \methodname \num{400} epoch model did not improve over the \num{200} epoch model.
In \Cref{fig:keypoints-dense-pose}, we report qualitative results for keypoint detection on randomly chosen images.

\begin{table}[tb]
   \centering
   \sisetup{
     table-format=2.1,
     round-mode = places,
     round-precision=1,
     detect-all,
  }
  \caption{Keypoint detection on COCO (R50-FPN).}
  \label{tab:keypoint-coco-eval}
  \footnotesize
  \begin{tabular}{lcSSS}
    \toprule
    Method & ep & AP$^{\text{kp}}$ & AP$^{\text{kp}}_{50}$ & AP$^{\text{kp}}_{75}$ \\
    \midrule
    Supervised & 100 & 65.3 & 87 & 71.3 \\
    Rand init  & {--} & 63 & 85.1 & 68.4  \\
    \midrule
    DenseCL~\citep{wang2021dense} & 200 & 66.3 & 87.1 & 71.9 \\
    PixPro~\citep{xie2021propagate} & 400 & 66.6 & 87.2 & 73.0 \\
    ReSim~\citep{wang2022cp} & 200 & 66.3 & 87.2 & 72.4 \\
    SetSim~\citep{wang2022exploring} & 200 & 66.7 & \bfseries 87.8 & 72.4 \\
    SlotCon~\citep{wen2022selfsupervised} & 200 & 66.5 & \uline{87.5} & 72.5 \\
    \midrule
    \methodname & 200 & \uline{66.9} & \uline{87.5} & \uline{73.2}\\
                & 400 & \bfseries 67.0 & 87.4 & \bfseries 73.3 \\
    \bottomrule
  \end{tabular}
\end{table}

\paragraph{Pascal VOC Object Detection.}

In \Cref{tab:voc-eval}, \methodname \num{200} epoch model performs comparably with PixPro~\citep{xie2021propagate}.
Similarly to keypoint detection, the \methodname \num{400} epoch model did not improve upon the \num{200} epoch version.

\begin{table}[tb]
   \centering
   \sisetup{
     table-format=2.1,
     round-mode = places,
     round-precision=1,
     detect-all,
  }
  \caption{Object detection on Pascal VOC (R50-C4).}
  \label{tab:voc-eval}
  \footnotesize
  \begin{tabular}{lcSSS}
    \toprule
    Method & ep & AP & $\textup{AP}_{50}$ & $\textup{AP}_{75}$ \\
    \midrule
    Supervised & 100 & 53.5 & 81.3  & 58.8 \\
    Rand init  & {--} & 33.8 & 60.2 & 33.1  \\
    \midrule
    DenseCL~\citep{wang2021dense} & 200 & 58.7 & 82.8 & 65.2 \\
    ReSim~\citep{wang2022cp} & 200 & 58.7 & 83.1 & 66.3 \\
    InsCon~\citep{yang2022inscon} & 200 & 59.1 & 83.6 & 66.6 \\
    PixPro~\citep{xie2021propagate} & 400 & \uline{60.0} & \bfseries 83.8 & \uline{67.7} \\
    cp2~\citep{wang2022cp} & 600 & 56.9 & 82.3 & 63.6 \\
    SlotCon~\citep{wen2022selfsupervised} & 200 & 57.3 & 82.9 & 64.3 \\
    SetSim~\citep{wang2022exploring} & 200 & 59.1 & 83.2 & 66.1 \\
    \midrule
    \methodname & 200 & \bfseries 60.1 & \uline{83.7} & \uline{67.7} \\
                & 400 & 59.9 & \bfseries 83.8 & \bfseries 67.8 \\
    \bottomrule
  \end{tabular}
\end{table}

\paragraph{COCO dense pose estimation.}

In \Cref{tab:dense-pose-coco-eval}, \methodname average performance beats supervised models trained on ResNet-50 and ResNet-100 backbones.
\Cref{fig:keypoints-dense-pose} shows \methodname's qualitative results for the dense-pose estimation downstream task.

\begin{table}[tb]
   \centering
   \sisetup{
     table-format=2.1,
     round-mode = places,
     round-precision=1,
     detect-all,
  }
  \caption{Dense pose estimation on COCO (R50-FPN).}
  \label{tab:dense-pose-coco-eval}
  \footnotesize
  \begin{tabular}{lcSSSS}
    \toprule
    Method & ep & AP$^{\text{bb}}$ & AP$^{\text{mb}}$ & AP$^{\text{gps}}$ & AP$^{\text{gpsm}}$ \\
    \midrule
    Supervised (R50)~\citep{wu2019detectron2} & 100 & 61.2 & 67.2 & 63.7 & 65.3 \\
    Supervised (R101)~\citep{wu2019detectron2} & 100 & 62.3 & 67.8 & 64.5 & 66.2 \\
    \midrule
    DenseCL~\citep{wang2021dense} & 200 & 63.0 & 67.7 & 65.7 & 66.7\\
    PixPro~\citep{xie2021propagate} & 400 & \uline{63.1} & \bfseries 68.3 & 66.2 & \uline{67.4}\\
    SlotCon~\citep{wen2022selfsupervised} & 200 & 62.8 & 67.4 & 65.3 & 66.4\\
    \midrule
    \methodname & 200 & \bfseries 63.2 & \uline{68.2} & \bfseries 66.6 & \bfseries 67.5 \\
    & 400 & \bfseries 63.2 & \bfseries 68.3 & \uline{66.3} & 67.3\\
    \bottomrule
  \end{tabular}
\end{table}

\paragraph{Notes on VICRegL.} VICRegL performance was surprisingly below expectations in many downstream tasks. 
While~\citet{bardes2022vicregl} reported AP of \textbf{59.5} for the same protocol and model (\texttt{resnet50\_alpha0p75.pth}) we used, our experiments resulted in AP of \textbf{27.6} on VOC07. Additionally, there is an open issue on VICRegL's official GitHub repo reporting the same reproducibility problem with similar results.

\section{Ablations}

To ablate the main hyperparameters of our model, we pre-trained \methodname on the ImageNet-1M dataset for \num{50} epochs and reported average performance results (3 independent runs) on Pascal VOC07 object detection.

\subsection{Multi-crop and the choice of loss function}

In \Cref{tab:multi-crop-loss-function}, we explore two loss functions that could be used in~\methodname's learning framework: the non-contrastive $\ell_2$-norm dot product and the ranking margin loss~\eqref{eq:ctx-to-local-loss}.
Moreover, we evaluate the effect of multi-crop augmentation on both loss functions. 
The $\ell_2$-normalized dot product loss, proposed by~\citet{grill2020bootstrap} and used in PixPro~\citep{xie2021propagate}, performs well with two views. 
However, performance decreases when multi-crop is employed. 
On the other hand, the ranking loss performs well in both setups as it can extract extra performance from multi-crop augmentation.

\begin{table}[tb]
   \centering
   \sisetup{
     table-format=2.1,
     round-mode = places,
     round-precision=1,
     detect-all,
  }
  \caption{Contrastive \vs non-contrastive loss functions and the effect of multi-crop augmentation. 
  }
  \label{tab:multi-crop-loss-function}
  \footnotesize
  \begin{tabular}{lcSSS}
    \toprule
    Loss & multi-crop & AP & $\textup{AP}_{50}$ & $\textup{AP}_{75}$ \\
    \midrule
    $\ell_2$ & \xmark  & 58.6 & 82.8 & \bfseries 66.2 \\
             & \cmark & 58.3 & 82.9 & 65.3 \\
    Rank & \xmark & 58.5 & 82.8 & 65.6 \\
            & \cmark & \bfseries 58.8 & \bfseries 83.3 & 65.9 \\
    \bottomrule
  \end{tabular}
\end{table}

\subsection{Normalized multi-head self-attention}
\label{sec:normalized-mhsa}

We propose a variation of the MHSA layer employed in Vision Transformers~\citep{dosovitskiy2021an}. 
Specifically, we normalize queries and keys before computing the attention scores. 
By normalizing the vector's magnitudes, we constrain the similarity scores to \num{-1.0} and \num{1.0}, which, in practice, avoids training instabilities and improves downstream task performance, \cf \Cref{tab:nmhsa-vs-mhsa}. 

\begin{table}[tb]
   \centering
   \sisetup{
     table-format=2.1,
     round-mode = places,
     round-precision=1,
     detect-all,
  }
  \caption{Normalized multi-head self-attention (NMHSA) performs slightly better than regular MHSA.}
  \label{tab:nmhsa-vs-mhsa}
  \footnotesize
  \begin{tabular}{lSSS}
    \toprule
    Method & AP & $\textup{AP}_{50}$ & $\textup{AP}_{75}$ \\
    \midrule
    MHSA & 58.3 & 83.1 & 65.8 \\
    NMHSA & \bfseries 58.7 & \bfseries 83.3 & \bfseries 65.9 \\
    \bottomrule
  \end{tabular}
\end{table}

\begin{table}
   \centering
   \sisetup{
     table-format=2.1,
     round-mode = places,
     round-precision=1,
     detect-all,
  }
  \caption{Negative sampling strategies for contrastive learning.}
  \label{tab:negative-sampling}
  \footnotesize
  \begin{tabular}{lcSSS}
    \toprule
    Method & queue & AP & $\textup{AP}_{50}$ & $\textup{AP}_{75}$ \\
    \midrule
    Inter & \cmark & 57.4 & 82.5 & 63.6\\
    Inter (avg) & \cmark & 57.5 & 82.8 & 64.7\\
    Intra & \xmark & \bfseries 58.7 & \bfseries 83.3 & \bfseries 65.9\\
    \bottomrule
  \end{tabular}
\end{table}

\subsection{Bootstrapping self-supervised signals}
\label{sec:positive-theshold}

To match local representations across different views of an image, we explore the spatial locality inductive bias present in natural images and expand it to the feature space. 
Intuitively, if two distinct pixels lie within a distance threshold $T_{pos}$, we assume their representations encode similar information. 
In \Cref{tab:pixel-neighborhood-threhold}, we explore the effect of the distance threshold used to identify pixels as neighbors across different views. 
As shown, too small or too large values for $T_{pos}$ invalidates the inductive bias assumption and harms the learned representations, \cf \Cref{fig:overview}.

\begin{table}
   \centering
   \sisetup{
     table-format=2.1,
     round-mode = places,
     round-precision=1,
     detect-all,
  }
  \caption{The effect of $T_{pos}$ on the learned representations.}
  \label{tab:pixel-neighborhood-threhold}
  \footnotesize
  \begin{tabular}{lSSSSS}
    \toprule
     & 0.5 & 0.6 & 0.7 & 0.8 & 0.9 \\
    \midrule
    $T_{pos}$ & 57 & 58.3 & \bfseries 58.5 & 58.1 & 57.7 \\ 
    \bottomrule
  \end{tabular}
\end{table}

\subsection{Exploring negative sampling strategies}
\label{sec:negative-sampling-strategies}

In \Cref{tab:negative-sampling}, we explore three negative sampling strategies for \methodname's loss function~\eqref{eq:ctx-to-local-loss}. 
For two strategies, we utilize an extra queue containing \num{16384} representations as a source of negatives. 
In the first strategy (inter), at each training iteration, we randomly take one local representation from the output feature map of the teacher branch and store it in the queue.
Older representations in the queue are discarded in favor of new ones. 
This way, the queue holds local representations from multiple images.
In the second strategy (inter avg), we aggregate the feature map into a single vector using a global average operator. 
Lastly, we use the local features without positive matchings from within the view as negatives. 
Since this strategy does not require negatives from other images (no queue), we call it intra-negative.
As shown in \Cref{tab:negative-sampling}, the intra-negative strategy outperforms the other ones in VOC07 and is~\methodname's default strategy.

\section{Implementation details}
\label{sec:impl-details}

We use the ResNet-50~\citep{he2016deep} architecture without the last fully connected and global average pooling layers as the feature extractor.
Following, the projection head is a two-layer MLP with \num{4096} hidden units, ReLU, batch normalization, and an output dimension of \num{256}.
To create views, we follow~\citepos{grill2020bootstrap} protocol.

We forward an image view $x \in \mathbb{R}^{3 \times 224 \times 224}$ and obtain a feature map $F \in \mathbb{R}^{256 \times 7 \times 7}$.
The contextualized prediction head $q_s$ implements the Normalized Multi-Head Self-Attention layer. 
It receives the feature map as input and trains \num{8} parallel attention heads. 
Each attention head learns independent query, key, and value matrices, $W^q, W^k, W^v \in \mathbb{R}^{256 \times 32}$.
To compute the attention scores, we normalize the projected queries and keys to unit vectors. 
The output of each head is concatenated (in the feature dimension) and passed through a linear output layer whose output has the same shape as the input.
 
\methodname is trained using \num{4} NVIDIA A100 GPUs, a total batch size of \num{2048} images, using the LARS~\citep{you2017large} optimizer, weight decay of \num{2e-5} and learning rate of \num{1.0} with a cosine decay schedule. In practice, the margin value in~\eqref{eq:ctx-to-local-loss} is set to $\mu=100$.

\subsection{Evaluation protocols}

\textbf{COCO detection and instance segmentation.} We followed the protocol from~\citet{he2020momentum} and fine-tuned all layers of a Mask-RCNN~\citep{he2017mask} on the \texttt{train2017} set ($\sim$118k images) and evaluated on \texttt{val2017}, using the $1\times$ schedule ($\sim$12 epochs). 

\textbf{Cityscapes instance segmentation.} We followed the~\texttt{mask\_rcnn\_R\_50\_FPN.yaml} config file from \texttt{Detectron2}~\cite{wu2019detectron2}, without changes, and fine-tuned all layers of a Mask-RCNN (R50-FPN backbone) for \num{24}k iterations, with a global batch size of \num{32} images (\num{8} per GPU), and a learning rate of \num{0.01}.

\textbf{LVIS object detection and instance segmentation.} We followed the \texttt{mask\_rcnn\_R\_50\_FPN\_1x.yaml} config file for LVISv1 instance segmentation from \texttt{Detectron2}, with no BN, and fine-tuned a Mask R-CNN (R50-FPN) on \texttt{lvis\_v1\_train} for \num{180}k iterations ($1\times \text{schedule}$) with a batch size of 16 (4 images per GPU), a learning rate of \num{0.001} and evaluated on \texttt{lvis\_v1\_val}.

\textbf{COCO keypoint detection.} We used the keypoint implementation of Mask R-CNN (R50-FPN) from \texttt{Detectron2}, fined tuned on \texttt{keypoints\_coco\_2017\_train}, and evaluated on \texttt{keypoints\_coco\_2017\_val} for \num{90}k iterations ($1\times \text{schedule}$), a batch size of \num{16} (\num{4} images per batch), a learning rate of \num{0.02}, and with enabled BN.

\textbf{Pascal VOC Object Detection.} We followed~\citepos{he2020momentum} protocol and fine-tuned all layers of a Faster R-CNN~\citep{NIPS2015_14bfa6bb} (R50-C4) on \texttt{trainval07+12} ($\sim$16.5k images) for \num{24}k iterations and evaluated on \texttt{test2007}.

\textbf{COCO dense pose estimation.}
We followed the DensePose~\citep{wu2019detectron2} project from \texttt{Detectron2} and fine-tuned a Faster R-CNN (R50-FPN) backbone using \methodname's pre-trained representations (1 $\times$ schedule).
Specifically, we used the \texttt{densepose\_rcnn\_R\_50\_FPN\_s1x.yaml} config file from the \texttt{Detectron2} repository, with BN enabled.

\section{Conclusions}

We presented \textbf{C}ontextualized \textbf{Lo}cal \textbf{V}isual \textbf{E}mbeddings (\methodname), a self-supervised method designed to learn representations to solve dense prediction tasks.
\methodname combines the multi-head self-attention layer commonly used in the Transformer model with convolutional backbones to learn prediction vectors that combine multiple similar areas of a view into a contextualized vector used to predict a local part of another view.
We empirically validate our design choices through a detailed ablative study of \methodname's main hyperparameters. 
Additionally, we extensively benchmarked \methodname in many downstream dense prediction tasks such as object detection, instance segmentation, keypoint detection, and dense pose estimation.
\methodname pre-trained representations showed robust performance against state-of-the-art SSL methods and supervised baselines. 

\section*{Acknowledgements}

The computations were performed in part on resources provided by Sigma2---the National Infrastructure for High Performance Computing and Data Storage in Norway---through Project~NN8104K.
This work was funded in part by the Research Council of Norway, through its Centre for Research-based Innovation funding scheme (grant no.~309439), and Consortium Partners.

This study was financed in part by the Coordenação de Aperfeiçoamento de Pessoal de Nível Superior---Brasil (CAPES)---Finance Code 001



{\small
\bibliographystyle{ieee_fullname}
\bibliography{abrv,egbib}
}

\end{document}